\documentclass[letterpaper]{article} 
\usepackage[arxiv]{aaai23}  
\usepackage{times}  
\usepackage{helvet}  
\usepackage{courier}  
\usepackage[hyphens]{url}  
\usepackage{graphicx} 
\usepackage{natbib}  
\usepackage{caption} 
\usepackage[frozencache,cachedir=.]{minted}
\usepackage{mathabx}
\usepackage[svgnames]{xcolor}
\definecolor{mgreen}{rgb}{0.6, 1.0, 0.6}
\definecolor{mred}{rgb}{1.0, 0.75, 0.79}
\definecolor{mblue}{rgb}{0.68, 0.93, 0.93}

\title{Synergistic Integration of Large Language Models and Cognitive Architectures for Robust AI: An Exploratory Analysis}

\author {
    Oscar J. Romero, 
    John Zimmerman, 
    Aaron Steinfeld, 
    Anthony Tomasic 
}
\affiliations {
    Carnegie Mellon University\\    
    oscarr@andrew.cmu.edu,  
    johnz@andrew.cmu.edu,
    steinfeld@cmu.edu, 
    tomasic@andrew.cmu.edu
}

\usepackage{bibentry}

\begin{document}

\maketitle

\begin{abstract}



This paper explores the integration of two AI subdisciplines employed in the development of artificial agents that exhibit intelligent behavior: Large Language Models (LLMs) and Cognitive Architectures (CAs). We present three integration approaches, each grounded in theoretical models and supported by preliminary empirical evidence. The modular approach, which introduces four models with varying degrees of integration, makes use of chain-of-thought prompting, and draws inspiration from augmented LLMs, the Common Model of Cognition, and the simulation theory of cognition. The agency approach, motivated by the Society of Mind theory and the LIDA cognitive architecture, proposes the formation of agent collections that interact at micro and macro cognitive levels, driven by either LLMs or symbolic components. The neuro-symbolic approach, which takes inspiration from the CLARION cognitive architecture, proposes a model where bottom-up learning extracts symbolic representations from an LLM layer and top-down guidance utilizes symbolic representations to direct prompt engineering in the LLM layer. These approaches aim to harness the strengths of both LLMs and CAs, while mitigating their weaknesses, thereby advancing the development of more robust AI systems. We discuss the tradeoffs and challenges associated with each approach.

\end{abstract}

\section{Introduction}

Pre-trained Large Language Models (LLMs) like ChatGPT, GPT-4, and PaLM 2 are generative models that excel in a variety of natural language tasks \cite{brown:2020, devlin:2019} and even show promise in interactive decision-making \cite{li:2022}, reasoning \cite{diao:2023, xie:2023, yao:2023}, and modeling aspects of artificial general intelligence (AGI) \cite{kosinski:2023, bubeck:2023}. However, LLMs face interpretability, consistency, and scalability issues \cite{mialon:2023}, partly due to limitations in context window size and sensitivity to prompt structure as they often rely on precise and carefully engineered instructions \cite{wei:2023}. They're criticized for being stochastic parrots and lacking detailed reasoning explanations \cite{bender:2021}. Hallucinations \cite{welleck:2019,qian:2022,wei:2023} and biases \cite{weidinger:2022, venkit:2022} are further concerns, affecting trustworthiness and ethical aspects \cite{huang:2023}. The dependence on larger models for better performance raises resource challenges \cite{mialon:2023}, and scalable LLMs incorporating continual learning are still an open question \cite{scialom:2022}.

In contrast, Cognitive Architectures (CAs) propose hypotheses about the fixed structures governing the operation of minds, whether in natural or artificial systems, facilitating intelligent behavior in complex environments \cite{llr:2017}. CAs like ACT-R \cite{anderson:2014}, SOAR \cite{laird:2019}, CLARION \cite{sun:2016}, and LIDA \cite{franklin:2006} model various human cognitive aspects: memory, learning, reasoning, perceptual-motor interaction, theory of mind, AGI, and more \cite{kotseruba:2020}. CAs prioritize bounded rationality, striving for satisfactory decisions under resource constraints, diverging from LLMs' pursuit of optimality. However, CAs face challenges in knowledge representation and scalability. Their encoded information is limited in size and homogeneous typology, meaning the knowledge processed by a cognitive agent\footnote{Hereafter, consider a cognitive agent as an artificial agent constructed on a particular CA.} is typically tailored for specific domains and tasks \cite{lieto:2018}. 

Unlike humans, CAs struggle with complex knowledge and their actions are confined to manually curated procedural knowledge \cite{park:2023}. According to \cite{marcus:2020}, LLMs struggle to derive cognitive models from discourse and lack capabilities to reason over those cognitive models\footnote{A cognitive model should at least include information about the entities in the external world, their properties, and their relationships with other entities, as well as the modeling of the cognitive processes that operate over those entities \cite{marcus:2020}.}. Hence, CAs could play a pivotal role in either augmenting or leveraging LLMs by contributing to the creation and dynamic updating of cognitive models. Likewise, cognitive models could be leveraged to better interpret LLMs' black-box learning algorithms and decision-making processes \cite{binz:2023}.

Both LLMs and CAs have made valuable and sound contributions to the construction of complex autonomous AI agents; however, each approach has its strengths and weaknesses (as summarized on Table \ref{tab:comparison}). Thus, the main contribution of this work lies in characterizing the plausible approaches to integrating CAs and LLMs, viewing them through a hybrid and synergetic lens.

\begin{table}[t]
\centering
\begin{tabular}{l|l|l}
    \textbf{Feature}                &  \textbf{LLMs} &  \textbf{CAs} \\ 
    Language processing              & ++    & -+ \\
    World knowledge                  & ++    & -+ \\
    Reasoning                        & -+    & ++ \\
    Symbolic processing              & -+    & ++ \\
    Connectionist processing         & ++    & -+ \\
    Knowledge scalability            & +-    & -+ \\
    Planning                         & -+    & +- \\           
    Learning                         & --    & +- \\ 
    Memory management                & --    & ++ \\
    Consistency (no hallucinations)  & -+ & ++ \\       
\end{tabular}
\caption{Feature comparison between LLMs and CAs. (++) Fully supported. (+-) Almost always supported. (-+) Sometimes supported. (--) Rarely (or not) supported.}
\label{tab:comparison}
\end{table}

\section{Relevant Work}

\textbf{Chain-of-thought prompting (CoT):} CoT prompting \cite{mialon:2023, diao:2023} enhances LLM reasoning, leading to improved performance in various reasoning and natural language processing tasks. CoT breaks down multi-step problems into intermediate steps, enabling the model to address reasoning problems. ReAct \cite{yao:2023} combines both reasoning (CoT prompts) and action (action plan generation). It organizes a workflow that decomposes task goals, injects task-relevant knowledge, extracts important observation components, and refines action plans based on feedback. Auto-CoT \cite{zhang:2022} proposes a model that samples questions with diversity and automatically generates demonstrations to correct mistakes in reasoning chains. The approaches we propose in this paper assume using CoT for problem decomposition, allowing a CA to inject its output into each reasoning step.

\textbf{Augmented Language Models}: it combines enhanced reasoning skills of an LLM with tools like APIs, DBs, and code interpreters for improved knowledge retrieval, reasoning, and action execution \cite{mialon:2023}. Program-Aided Language model (PAL) \cite{gao:2023} reads natural language problems, generates intermediate programs for reasoning, and delegates the solution step to a Python interpreter. Toolformer \cite{schick:2023} is a model trained to decide which APIs to call, when to call them, what arguments to pass, and how to best incorporate the results into future token prediction. Our modular approach extends the idea of augmenting an LLM with cognitive processing and assumes the usage of external APIs. 

\textbf{CAs and LLMs}: Generative Agents \cite{park:2023} is a model that uses a cognitive architecture and an LLM to generate realistic behavior. It defines three components: a memory stream for recording comprehensive experiences in natural language, a reflection component for deriving higher-level inferences about self and others, and a planning component translating these inferences into action plans. This approach differs from ours in that it does not use symbolic structures but unstructured natural language. OlaGPT \cite{xie:2023} is an LLM cognition framework aiming to solve reasoning problems with human-like problem-solving abilities by leveraging CoT. OlaGPT proposes to approximate cognitive modules, such as attention, memory, learning, reasoning, action selection, and decision-making. The first case of our modular approach resembles OlaGPT to some extent.

Open-source experimental applications like Auto-GPT\footnote{\url{https://github.com/Significant-Gravitas/Auto-GPT}} and BabyAGI\footnote{\url{https://github.com/yoheinakajima/babyagi}} aim to advance AGI. Auto-GPT manages long-term and short-term memory, language generation, and summarization. BabyAGI uses LLM chains to perform tasks based on goals. These approaches hold significant potential and are likely to integrate further with human cognition modeling. Although with not a strict commitment to model a cognitive architecture, Voyager \cite{wang:2023} facilitates continual learning through an evolving code library for complex behaviors. An iterative prompting mechanism incorporates feedback, errors, and self-verification for program improvement. 
\cite{lecun:2022} outlines the considerations for crafting a cognitive architecture using energy minimization mechanisms, enabling reasoning, prediction, and multi-scale planning. They emphasize that while deterministic generative architectures withstand energy distribution issues, non-deterministic structures like auto-encoders and joint embeddings are susceptible to collapse.

\section{Integration Approaches}


In this section, we propose and discuss the tradeoffs of three different approaches for the integration of CAs and LLMs: the modular approach, the agency approach, and the neuro-symbolic approach. To illustrate the practical implementation of each approach, we base our examples on a scenario involving a cognitive agent designed to assist people with visual impairments in everyday tasks such as navigation and exploration of indoor environments, effective use of public transportation, etc. The agent operates on a smartphone device, utilizing sensor data processing, computer vision for object detection, and speech recognition to perceive its environment. Its actions encompass language generation and invocation of external APIs. The agent engages in conversation with its user, reasons about their needs and requests, constructs shared mental models to achieve goals effectively, and makes decisions that unfold in the short and long term. 

For the remainder of this paper, let us consider that the inputs of an LLM can be multimodal, involving text and images, while the outputs are exclusively text-based. Conversely, for the sake of simplicity, CAs' inputs and outputs are limited to formatted text, although, in practice, various CAs can process diverse modalities. As a reference framework for CAs' structure, our approach adopts the Common Model of Cognition (CMC) \cite{llr:2017}, which captures a consensus regarding the structures and processes that resemble those found in human cognition. CMC defines five high-level modules, including perception, motor, working memory, declarative long-term memory, and procedural long-term memory, each of which can be further decomposed into multiple sub-modules. Behavior in the CMC is organized around a cognitive cycle driven by procedural memory, with complex behavior (e.g., reasoning, planning, etc.) emerging as sequences of such cycles. In each cognitive cycle, the system senses the current situation, interprets it with respect to ongoing goals, and then selects an internal or external action in response. Both the agency and the neuro-symbolic approaches use different reference frames, which will be discussed later.

\subsubsection{Modular Approach}

\begin{figure*}[tbh]
\centering
\includegraphics[width=0.9\linewidth]{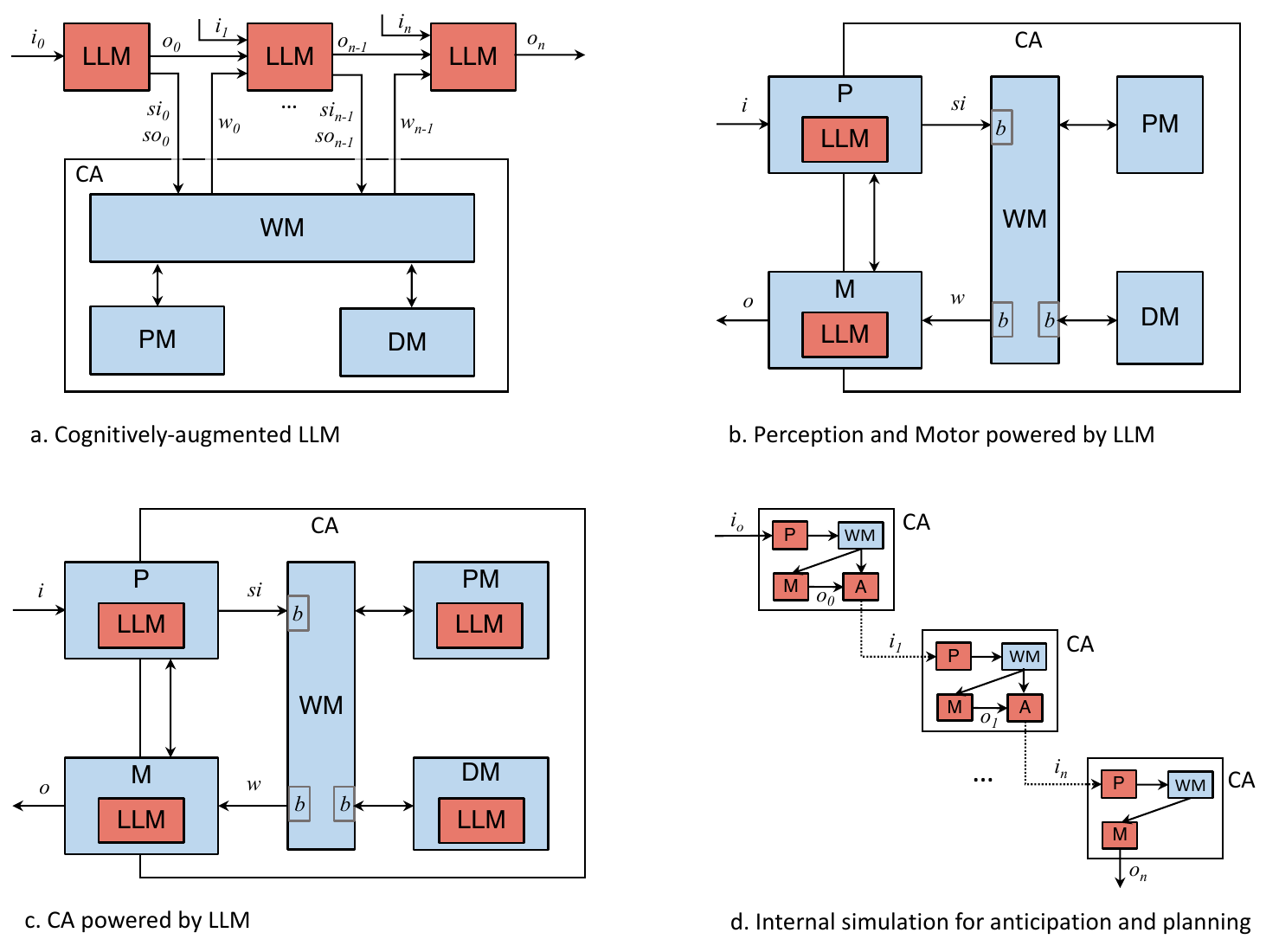} 
\caption{Modular approach. (a) Chain-of-Thought or recursive reasoning that augments an LLM with content generated by a CA. (b) Perception and Motor modules of a CA that leverages the power of LLMs. (c) Multiple modules of a CA that use LLMs to process and/or retrieve data. (d) A CA that leverages LLMs to predict/anticipate future states of the environment in order to perform reasoning and planning (some modules are not shown for the sake of legibility). Red-colored boxes denote LLMs and blue-colored ones denote CAs modules. Perception (P), motor (M), working memory (WM), long-term procedural memory (PM), long-term declarative memory (DM), and Anticipation (A) correspond to modules of a CA. $i$ and $o$ correspond to the input and output of the system, respectively. $si$ and $so$ are symbolic representations of the input $i$ and the output $o$, respectively. $w$ corresponds to the contents of the working memory. $b$ are module-specific working memory buffers. Solid arrows denote the flow of information and dotted arrows denote predictions of the next input.}
\vspace{-0.5cm}
\label{fig:modular}
\end{figure*}

A straightforward way to integrate LLMs and CAs is using a modular approach where either (1) LLMs partially enhance the performance of certain modules and components of a CA, or (2) a CA augments an LLM by injecting reasoning traces and contents from memories into the prompting process. Figure \ref{fig:modular} depicts 4 different cases of modular integration. This integration allows modules to be easily replaced by LLMs or their CA module counterparts. 

Case (a) assumes a recursive prompting scenario \cite{mialon:2023} where an LLM decomposes a complex problem into subproblems, and the intermediate outputs are aggregated to generate a final output. In this case, a CA could be used to prime every intermediate step at the LLM with reasoning traces from procedural knowledge as well as relevant content from memories. The mechanism would be as follows: given an initial input $i_0$ (e.g., a user's request, external signals, etc.), the LLM generates an intermediate output $o_0$ (e.g., the first step towards the solution of the user's request) and a set of equivalent symbolic structures for both the input, $si_0$ (e.g., intents, entities, and properties recognized from the input), and the output, $so_0$ (e.g., symbolic representation of LLM's actions and reasoning steps)\footnote{Empirical work demonstrates that LLMs can be trained and fine-tuned to learn to encode key aspects of traditional symbolic structures \cite{pavlick:2023, romero:2021, tomasic:2021}.}. The CA uses those symbolic structures as inputs and executes one or several cognitive cycles, after which, the contents of the working memory ($w_0$), including fired productions, relevant information from declarative memories, and actions, are injected as cues into the next intermediate step of the LLM. The process repeats until a final output is generated. 

Consider this streamlined example: A bus rider employs the term ``downtown'' ambiguously as the destination. Depending on the day of the week, the user may refer to two specific places in the downtown area, namely the workplace or the gym. The cognitive agent uses a combination of declarative and procedural knowledge to discern the user's transit patterns and effectively disambiguate the term downtown. The triggered productions and relevant contents of the working memory are subsequently integrated into the next LLM's recursive step, thereby guiding and enhancing its reasoning process with new information. 

% ...

\vspace{0.2cm}
\begin{minted}[fontsize=\scriptsize,escapeinside=||,mathescape=true]{text}
|\textbf{[i$_0$]}|  User: "when is my next bus to downtown coming?"
|\textbf{[o$_0$]}|  search bus schedule with destination |\colorbox{mgreen}{downtown}|
|\textbf{[si$_0$]}| (intent, (inform, destination, downtown))
|\textbf{[o$s_0$]}| (step, (search, orig, X, destination, downtown))
|\textbf{[w$_0$]}|  (semantic, (is_a (downtown, place)))
      (semantic, (today (saturday)))
      (episodic, (destination=downtown, place=workplace, 
           |$\drsh$| days=[monday... friday]))
      (episodic, (destination=downtown, place=gym, 
           |$\drsh$| days=[saturday, sunday]))
      (procedural, (if destination == x and today == y 
           |$\drsh$| then destination=place))
      (semantic, (destination (gym)))
|\textbf{[o$_1$]}|  search bus schedule with destination |\colorbox{mgreen}{gym}|
...
\end{minted}


Now, case (b) proposes an integration scenario where a CA leverages multimodal LLMs to support both the perception and motor modules of the system. At the perception level, the LLM converts external signals (e.g., text and images) into symbolic representations and their relations, along with relevant metadata, which is then stored in the corresponding working memory buffers. For instance, we used a description of an image generated by GPT-4\footnote{At the time of writing this paper, OpenAI is holding back GPT-4 image processing features, so we used a natural language description generated with GPT-4 and reported in \cite{zhu:2023}.}, then we provided an in-context learning example to ChatGPT to convert the description into symbolic structures:


\vspace{0.2cm}
\begin{minted}[fontsize=\scriptsize,escapeinside=||,mathescape=true]{text}
|\textbf{[prompt]}| your job is to transform that natural language 
    description into symbolic representations. Example, 
    |\colorbox{mgreen}{Input:}| "this is an office with a desk, a computer on 
    top of it, and a chair. There is also a printer on 
    the left of the computer and some pencils on the 
    right". |\colorbox{mgreen}{Output:}| has(office, desk). has(office,  
    computer). has(office, chair). onTopOf(computer, 
    desk). toTheRightOf(printer, computer). toTheLeftOf(
    computer, printer). toTheLeftOf(computer, pencils)..
|\textbf{[prompt]}| |\colorbox{mgreen}{Input:}| "a busy city street with a clock tower 
    in the background. The street is lined with shops 
    and restaurants on both sides. There are several 
    motorcycles parked on the side of the road, and 
    people are walking down the street. The clock tower 
    has a large clock face with roman numerals and a 
    small spire on top...". |\colorbox{mgreen}{Output:}|
|\textbf{[ChatGPT]}| has(city_street, clock_tower). has(clock_tower
    , clock_face). has(clock_face, roman_numerals). has(
    clock_face, small_spire). has(city_street, shops). 
    has(city_street, restaurants). has(city_street, 
    motorcycles). toTheLeftOf(clock_tower, city_street). 
    toTheRightOf(city_street, clock_tower). toTheRightOf
    (shops, city_street)... onTopOf(roman_numerals, 
    clock_face). onTopOf(small_spire, clock_face). 
    onTopOf(clock_face, clock_tower)...
\end{minted}
\vspace{0.2cm}

As observed in the example, though not entirely accurate, LLMs demonstrate the capability to extract high-level compositional and spatial relationships between entities from a given image/text and then re-express them using symbolic representations. After generating and storing these symbolic structures in the working memory, other modules of the CA can access them and perform diverse kinds of cognitive processes. Considering our initial example, it is expected that this symbolic representation of perceived images will enable both the visually impaired user and the cognitive agent to collaboratively construct shared mental models for navigation, thereby enhancing spatial cognition and situational awareness of the user. 
%
Conversely, the LLM-based motor module converts the symbol structures that have been stored in the working memory buffers into external actions (e.g., natural language generation, motor control, etc.)

Unlike case (b), which loosely integrates LLMs and CAs, case (c) proposes an integration where not only the perception/motor modules are driven by LLMs, but also the procedural and declarative (semantic and episodic) memories. Prior research \cite{park:2023} suggested using LLMs to retain episodic knowledge as lists of observations (depicting agents' behaviors in natural language). These can be synthesized into high-level observations using LLMs' summarization abilities, enabling agents to reflect on their experiences across different time spans. From another perspective, we envision the possibility of converting these natural language descriptions into symbolic structures using a proper method for fine-tuning or prompt-engineering an LLM.

Similarly, the large amount of factual knowledge directly accessible through LLMs can be harnessed to automatically extract knowledge and populate a semantic memory (e.g., an ontology) of a CA, which would otherwise require laborious manual knowledge curation and acquisition. Preliminary work in this area aims to populate arbitrary complex nested semantic knowledge schemas \cite{caufield:2023}. 

Regarding the procedural module, previous work \cite{park:2023} has shown that LLMs can be leveraged to create detailed step-by-step plans by translating the agent's observations of the environment expressed in natural language into high-level action plans, which are then recursively decomposed into detailed behaviors for action and reaction. Alternatively, we anticipate that, with the proper in-context prompting or fine-tuning, an LLM can translate natural language descriptions into symbolic structures such as production rules. For instance, consider the example below where we prompted ChatGPT to generate an ACT-R production rule for filtering out transit searches based on a natural language description and a single example (borrowed from \cite{actr:2023}) used to prime the LLM\footnote{The complete log is available here: \url{https://shareg.pt/nO1zssm}.}:


\begin{minted}[fontsize=\scriptsize,escapeinside=||,mathescape=true]{text}
|\textbf{[prompt]}| |\colorbox{mgreen}{Instructions:}| ... your job is to generate ACT-R
    productions based on a natural language description 
    that I will provide to you... For instance: 
    |\colorbox{mgreen}{Description:}| generate an ACT-R production that 
    counts until a given number. |\colorbox{mgreen}{Production:}|
      (p example-counting
         =goal>....
|\textbf{[prompt]}| |\colorbox{mgreen}{Description:}| generate an ACT-R production such 
    that if I need to take a bus, check first the 
    weather and the traffic patterns. If it is snowing 
    and the traffic is heavy, then search only for bus 
    routes without transfers and minimum walking 
    distance. |\colorbox{mgreen}{Production:}|
|\textbf{[ChatGPT]}|
    (p check-weather-traffic-snow-heavy
       =goal>
         isa         take-bus
         state       check-weather-traffic
      =retrieval>                        
         isa         weather-traffic
         weather     snowing         
         traffic     heavy           
   ==>                                  
      =goal>                            
         isa         take-bus        
         state       search-routes   
         |\colorbox{mred}{\em{transfer    none}}|
         |\colorbox{mred}{\em{walking     min-distance}}|         
      +retrieval>                       
         isa         bus-routes      
         transfer    none            
         walking     min-distance)
\end{minted}
\vspace{0.2cm}

The condition part of the production checks both that the state of the goal is {\small \verb|check-weather-traffic|} and that the retrieved declarative chunk has the values {\small \verb|snowing|} and {\small \verb|heavy|} for the slots {\small \verb|weather|} and {\small \verb|traffic|}, respectively. If so, the action will change the goal's state to {\small \verb|search-routes|} and retrieve a declarative chunk for bus routes with no transfers and minimum walking distance. 

Although the generated production captures correctly the intention of the natural language description, it contains redundant slots for {\small \verb|transfer|} and {\small \verb|walking|} on the goal buffer of the action part (in italics). This type of transformation from natural language descriptions to symbolic productions can allow users to instruct and teach explicit procedural knowledge to their agents via conversation and, more interestingly, transform symbolic productions back into natural language for explainability purposes. However, it is not clear how an LLM could keep consistency between learned productions for a large knowledge base. Additionally, at least at its current state, LLMs by themselves cannot compile certain operations over the procedural memory such as conflict resolution and execution, so an LLM would still require an external interaction with a CA's procedural engine.


Finally, case (d) presents a streamlined approach to the simulation theory of cognition, which states that cognitive functions like planning and anticipation stem from internally simulated interactions with the environment \cite{shanahan:2006,hesslow:2012}. By inputting appropriate contextual information (such as working memory contents, sensory input, motor responses, and past experiences), we postulate that LLMs have the potential to forecast likely representations of the world's states resulting from the current state. That is, upon receiving an initial sensory input ($i_0$), the CA progresses through its standard perception-action pathway. Subsequently, rather than executing the resulting action ($O_0$) in the real world, the action $O_0$, along with the working memory contents, are used as inputs of the LLM. The LLM then generates a prediction for the next world state ($i_1$), which serves as a simulated input. Next, a simulated output {$o_1$} is generated, and then the process cycles until a certain condition is met. By chaining these sequences of perception-action-prediction, the system could anticipate the outcomes of its actions across multiple temporal scales. 

These simulated sequences may take on either linear configurations, predicting only one world state per simulated step, or branching tree-like structures, predicting multiple likely world states per step (tangentially related work has been proposed by \cite{yao:2023tree}). In the latter case, a planning mechanism could explore different branches of the tree by assessing their likelihood of occurrence and performing backtracking when necessary. As proposed by \cite{park:2023}, an LLM can be prompted to rank its answers based on a certain scale, similarly, we can prompt it to ``guesstimate'' a probability for each node of the tree. 

Below is a simplified example where the inputs to the LLM are the contents of the working memory (green), simulated user actions (red), and simulated system actions (blue). For simplicity, both user and system actions are natural language and do not involve symbols:


\vspace{0.2cm}
\begin{minted}[fontsize=\scriptsize,escapeinside=||,mathescape=true]{text}
|\textbf{[prompt]}| You are an intelligent agent that assists 
    |\colorbox{mgreen}{a person who is blind}| in retrieving information 
    from public transportation. |\colorbox{mgreen}{Today is snowing}| and 
    therefore the |\colorbox{mgreen}{traffic is heavy}|. Predict 3 things 
    that the user will ask you to do and assign a 
    probability to occur to each one.
|\textbf{[ChatGPT]}| |\colorbox{mred}{Check current bus/train delay}| (0.6), 
    |\colorbox{mred}{suggest alternative routes}| (0.3), 
    |\colorbox{mred}{provide weather-related alerts}| (0.1)
|\textbf{[prompt]}| The user requests you to provide weather-related 
    alerts, and |\colorbox{mblue}{you provide those alerts}|. What do you 
    predict the user will ask next?
|\textbf{[ChatGPT]}| |\colorbox{mred}{Ask for specific service disruptions}| (0.4), 
    |\colorbox{mred}{request tips for navigating in snowy conditions}| 
    (0.3), |\colorbox{mred}{inquire about expected clearing times}| (0.2)
\end{minted}

\subsection{Agency Approach}

\begin{figure*}[tbh]
\centering
\includegraphics[width=0.9\linewidth]{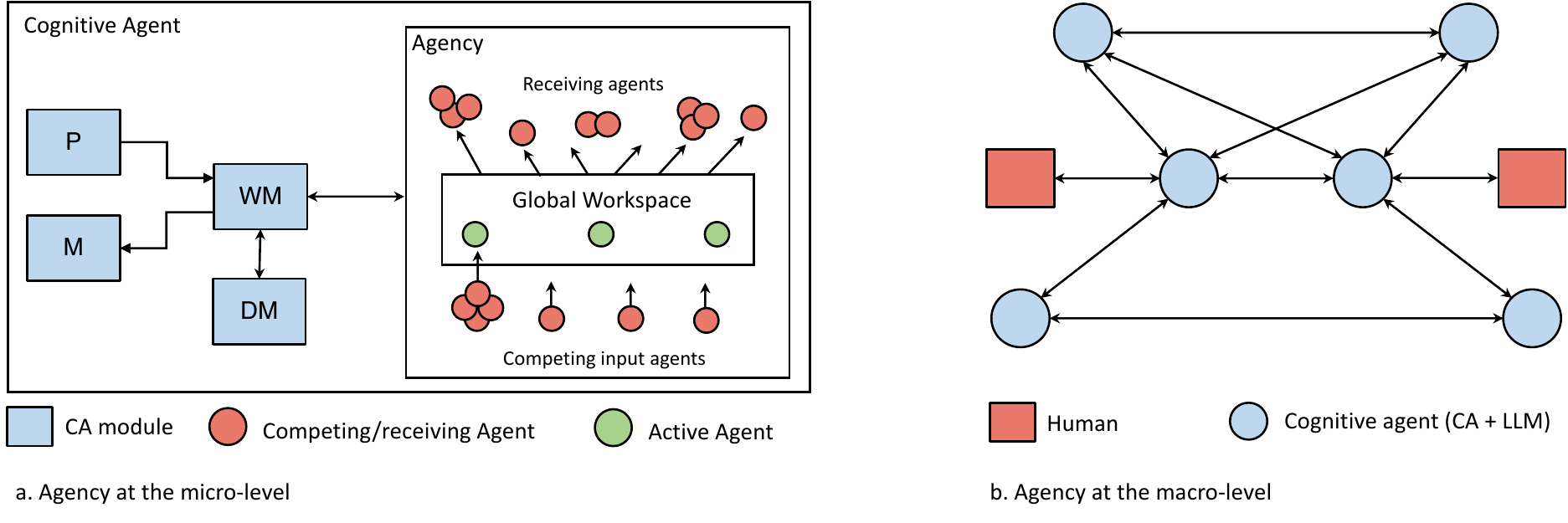} 
\vspace{-0.2cm}
\caption{Agency approach. a) Agents at the micro-level compete for resources and cooperate in decision-making. b) Agents at the macro-level interact with other agents and humans to cooperate in task resolution. P (Perception), M (Motor), WM (working memory), and DM (declarative memory) are modules of a CA.}
\vspace{-0.2cm}
\label{fig:agency}
\end{figure*}

The Agency approach operates on two levels - micro and macro (see Figure \ref{fig:agency}). Inspired by the Society of Mind theory \cite{minsky:1988} and LIDA cognitive architecture \cite{franklin:2006}, micro-level agency occurs within the cognitive architecture itself. Specialized agents process information in parallel, competing for resources like attention and memory. They collaborate by forming coalitions for decision-making and problem-solving. In contrast, macro-level agency involves cognitive agents interacting with other agents and humans to collaboratively achieve goals.

Consider the case of our cognitive agent designed to aid blind users in indoor navigation. At a micro-level, each agent operates through either a fine-tuned LLM or a symbolic processor. Cognitive processing unfolds as follows: sensory inputs are processed by the perception module, yielding abstract entities like objects, categories, actions, events, etc., forwarded to the working memory. Then, the working memory cues declarative memories to establish local associations, e.g., user navigation preferences, place familiarity, and more. Specialized agents at the agency observe working memory contents and form coalitions. 

For instance, object detection and semantic localization constitute one coalition, while natural language understanding and semantic grounding form another. These coalitions are transferred to the Global Workspace, where a competitive process selects the most relevant coalition. If a user approaches a staircase lacking a handrail, the coalition involving object detection and semantic localization takes precedence, globally transmitting its contents (e.g., staircase proximity and orientation) to other agents. In subsequent cognitive cycles, the coalition for natural language generation would be chosen to provide timely warnings to the user. 

While not a novel architectural approach, its potential lies in the diverse roles agents can assume within coalitions. For instance, an LLM agent engages in pair work, processing text or images to produce symbols, while a symbolic agent infers insights from these symbols. Another scenario involves one LLM agent fine-tuned to convert symbol structures into natural language text and another serving a supervisory role, pinpointing errors in the first agent's output.


Now, to better understand macro-level interactions, let's consider two users (\emph{A} and \emph{B}) alongside their cognitive agents (\emph{a} and \emph{b}). Agents \emph{a} and \emph{b} collaborate to exchange knowledge and intentions (e.g., \emph{a} shares spatial insights with \emph{b} of previous \emph{A}'s exploration of a building, thus aiding \emph{B}'s future navigation), negotiate (e.g., \emph{a} and \emph{b} helping teammates \emph{A} and \emph{B} reach an agreement when having conflicting goals), debate (e.g., \emph{a} and \emph{b} debating about their reasoning processes to approach a problem while reaching a consensus \cite{du:2023}), among others. All these kinds of interactions among agents could use natural language in order to foster transparency and interpretability, from the user's point of view, of the reasoning processes and conciliated actions, although the necessity of symbolic counterparts remains unclear.

\subsection{Neuro-Symbolic Approach}

\begin{figure*}[tbh]
\centering
\includegraphics[width=\linewidth]{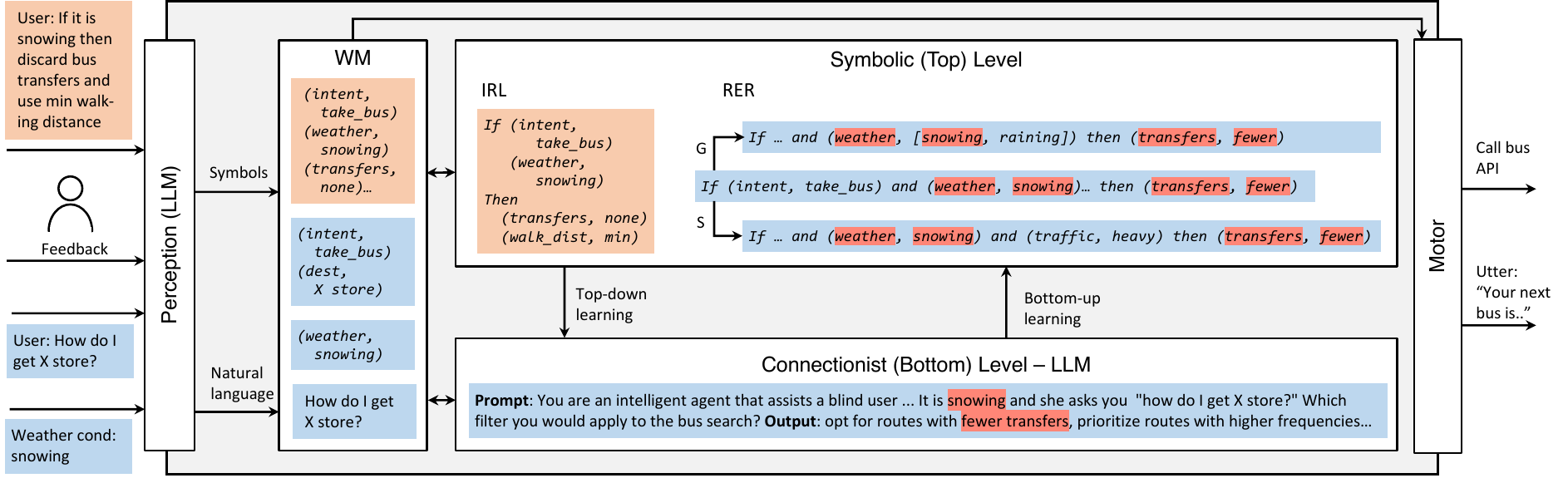} 
\caption{Neuro-symbolic approach. WM: Working Memory. IRL: Independent Rule Learning. RER: Rule Extraction Refinement. G: Generalization scenario. S: Specialization scenario. Orange-colored boxes illustrate the IRL case while the blue-colored boxes illustrate the RER case. Highlighted text represents entities and keywords present at the bottom level that are further extracted and translated into symbols at the top level.}
\vspace{-0.3cm}
\label{fig:neurosymbolic}
\end{figure*}

We present a neuro-symbolic approach inspired by the CLARION cognitive architecture, focusing primarily on the action-centered sub-system (ACS), while acknowledging the existence of three additional sub-systems within the architecture. The ACS operates across two distinct levels: the top level (symbolic), responsible for encoding explicit knowledge, and the bottom level (connectionist), tasked with encoding implicit knowledge. Consequently, the architecture exhibits a degree of redundancy in knowledge representation. These two levels synergistically engage in action selection, reasoning, and learning processes. Our focus is to explore the incorporation of LLMs at the bottom level, enhancing the knowledge extraction and integration process while exhibiting potential scalability towards novel scenarios. Further details on the mathematical model underpinning the cognitive processes can be found in \cite{sun:2016}.

CLARION defines three types of symbolic rules at the top level. The fixed rules (FR) are rules that have been hardwired by an expert and cannot be deleted; Independent-Rule-Learning (IRL) rules are independently generated at the top level, with little involvement (or no involvement at all) of the bottom level, which can be refined or deleted as needed; and Rule-Extraction-Refinement (RER) rules which are extracted from the bottom level. Figure \ref{fig:neurosymbolic} illustrates the process wherein a human provides a natural language instruction to create a new rule and the LLM-based perception module extracts symbolic structures that are further stored in the working memory. Through a template-matching mechanism, the contents of the working memory are expressed as an IRL rule where both its condition and action parts are chunks composed of dimension-value pairs\footnote{Each dimension may have one or multiple values associated.}, e.g., \emph{chunk$_i$((intent, take\_bus), (weather, snowing)) $\rightarrow$ chunk$_j$((transfers, none), (walk\_distance, min))}. 

On the other hand, if an action determined at the bottom level proves successful (according to a certain criterion), an RER rule is formulated and subsequently incorporated into the top level, e.g., given the output generated by the LLM at the bottom level\footnote{See full output log here: \url{https://sharegpt.com/c/LYIz9in}} on Figure \ref{fig:neurosymbolic}, the corresponding RER rule is \emph{chunk$_i$((intent, take\_bus), (weather, snowing)) $\rightarrow$ chunk$_j$((transfers, fewer))}. During subsequent interactions with the environment, the rule is refined based on the outcomes of its application: if the result is deemed successful, the rule's conditions may be generalized to make it more universal by adding new values to dimensions (e.g., \emph{chunk$_i$((intent, take\_bus), (weather, [snowing, \textbf{raining}])) $\rightarrow$ chunk$_j$((transfers, fewer))}). Conversely, if the outcome does not yield success, the rule should be specialized by removing values from dimensions or by adding new dimension-value pairs (e.g., \emph{chunk$_i$((intent, take\_bus), (weather, snowing), \textbf{(traffic, heavy)}) $\rightarrow$ chunk$_j$((transfers, fewer))}). 

Rule selection in IRL is determined by an information gain function, while RER uses a Boltzmann distribution based on rule's utility function and a base-level activation. The integration of both levels can be achieved through various mechanisms. Stochastic selection involves choosing a level (top or bottom) and a group of rules if the top level is chosen (e.g., FR, RER, or IRL). These selections are based on probabilities assigned by a metacognitive module to each level/group. Integration through bottom-up rectification occurs when the top level rectifies and incorporates outcomes from the bottom level (e.g., the LLM may discover additional dimension-value pairs not specified by the top level like \emph{``prioritize routes with higher frequencies''}). Alternatively, top-down guidance involves the bottom level utilizing outcomes from the top level, combined with its own knowledge, to make action decisions. This top-down guidance can be achieved by using prompt engineering techniques to prime the LLM with either FR or IRL rules.

Bottom-up learning is facilitated by the rule extraction mechanism, whereas top-down learning can be realized by using both FR and IRL rules as exemples to fine-tune the LLM at the bottom level. Determining whether an outcome from the bottom level is successful requires feedback, often in the form of rewards or reinforcement, which might not be readily available. To address this challenge, we propose two approaches: the incorporation of human-in-the-loop interactions, where feedback ensures the coherence of extracted rules, and the utilization of an additional LLM for self-play interactions emulating human feedback. Overall, both the bottom-up and the top-down learning mechanisms support explainability of decision-making and reasoning processes performed by the LLM at the bottom level. 

Harnessing LLMs at the bottom level of a CLARION-like architecture can contribute remarkably to enhancing the system's flexibility and scalability. First, unlike backpropagation neural networks used in CLARION, LLMs are not restricted to a fixed number of features and labels. Also, the LLMs-based variation we propose do not require to pre-define dimension-value pairs as CLARION does. Consequently, the utilization of LLMs at the bottom level can enable enhanced representational flexibility, with cascading benefits reaching the top level. Secondly, the conversion from unstructured natural language to symbols and vice versa can be executed seamlessly by an LLM-based bottom level. Lastly, leveraging an LLM with such broad knowledge of the world, coupled with cross-level learning dynamics and human feedback, can foster continuous learning loops where knowledge is constructed and refined over time.

\section{Discussion}

Among the three approaches discussed so far, there are some commonalities that we highlight next. First, the working memory, along with the perception module, plays an important role in retaining the most pertinent information while filtering out irrelevant stimuli. This contrasts with the idea of a context window in LLMs, where truncation strategies arbitrarily delete the oldest tokens observed when the length of the window reaches a maximum, potentially discarding critical parts of the context. The contents of the working memory are selectively and intentionally stored and recalled from long-term memories, allowing the agent to continuously interact with the environment without losing track of events. A second common aspect among all three approaches is the utilization of LLMs to accurately translate unstructured natural language to symbols and vice versa, as well as to extract factual knowledge about the world. This breakthrough opens up a realm of new possibilities, allowing for the seamless scaling of CAs to tackle complex real-world problems. 

Third, the three approaches can benefit from multi-modal multi-turn interaction. In cases where cognitive agents collaborate with humans, there is an opportunity to incrementally refine shared mental models of a task through continuous conversational interaction and scene understanding. Fourth, since all the approaches depend, in one way or another, on LLMs, they are susceptible to the stochastic nature of LLMs. This stochastic nature leads to variations (sometimes remarkable) in the outputs, even when the model is prompted with exactly the same input. And fifth, all three approaches contribute, to a greater or lesser extent, to the continuous construction of cognitive models about the entities in the world, their relationships, and the distinct cognitive processes that operate over them.

Regarding the Modular approach, the main difference among the four cases presented is the degree of integration between an LLM and a CA. The first case, the cognitively augmented LLM, aligns with the current trend of augmenting LLMs with external tools and interpreters and represents the most loosely integrated model among the four. In this case, the LLM retains control of execution, and the outputs of the CA are solely utilized for in-context learning purposes. The strength of this approach is that recursive LLMs receive gradual guidance during the chain-of-thought reasoning process. However, a notable disadvantage is that, due to the lack of overall control, the CA components can only contribute to reactive (System 1) responses rather than deliberative, high-order (System 2) ones.

The second case of the modular approach presents a moderately integrated model where only the perception and motor modules of a CA are powered with LLMs. The main strength of this model is that it aligns with the evident benefits obtained from multi-modal LLMs, which notably enhance text and image understanding, avoiding the need for task-specific and laborious labeling and training of machine learning models. Another advantage of this case is that it assumes a straightforward transformation from sensory inputs to symbolic percepts, which facilitates further processing. However, one main disadvantage is that the other modules of the CA still do not fully leverage the power of LLMs.

The third case presents a tightly integrated model that leverages the synergistic interaction between LLMs and symbolic components of a CA. LLMs extract factual knowledge from the world, automatically populating ontologies. These semantic representations then facilitate the creation of world models, addressing a limitation of LLMs. Furthermore, proper LLM's prompt engineering techniques would produce syntactically and semantically correct CA productions, which can be later compiled by a symbolic engine. However, a drawback of this integrated system is its heavy reliance on LLM outputs, rendering it susceptible to cascading failures, including hallucinations and biases.

The fourth case represents the most tightly integrated model. It involves a module designed for simulating the outcomes of future events. The primary advantage of this case is its capability to anticipate and plan by traversing and backtracking a tree-like structure of possible events. However, similar to the third case, this system heavily relies on the outputs of the LLM, which might occasionally be inconsistent. This inconsistency could lead to erroneous predictions in the early stages of internal simulation, resulting in cascading errors in the planning process.

Unlike the Modular approach, which can suffer from overall failures and inconsistencies if individual modules are poorly designed, the Agency approach at the micro-level offers greater robustness from two key angles. First, agents may encode redundant knowledge, resulting in multiple agents capable of achieving the same competence. This redundancy enhances system resilience as individual agents may fail, yet the system can still yield satisfactory outcomes. Second, agent role-playing strategies enable the system to self-reflect and promptly rectify potential deviations in reasoning processes. At the macro-level, the Agency approach stands out as the only one among the three approaches that considers inter-agent interactions, with a primary focus on collaborative interactions between agents and humans. However, aspects such as communication, coordination, hierarchies, etc. between agents still remain open questions.

The Neuro-symbolic approach is arguably the most tightly integrated model. It leverages the capabilities of LLMs to seamlessly translate unstructured natural language into structured symbolic representations and vice versa. This approach plays a crucial role in extracting rules from the connectionist level and subsequently generalizing and specializing those extracted rules over time. The interactions between the symbolic and connectionist levels enable the continuous construction of explainable models for decision-making and procedural processing based on black-boxed LLMs. However, a potential weakness of this approach lies in its heavy reliance on the LLM layer.


\section{Conclusions}

In this paper, we present three different approaches to integrating Cognitive Architectures and Large Language Models from an architectural perspective: a modular approach, an agency approach, and a neuro-symbolic approach. We discuss the trade-offs associated with each approach and provide insights for future research in this area.

\section*{Acknowledgements}

The contents of this paper were developed under grants from the National Institute on Disability, Independent Living, and Rehabilitation Research (NIDILRR grant numbers 90DPGE0003 and 90REGE0007)

\bibliography{bibliography}

\begin{thebibliography}{43}
\providecommand{\natexlab}[1]{#1}

\bibitem[{{ACT-R Website.}(2015)}]{actr:2023}
{ACT-R Website.} 2015.
\newblock Unit 1: Understanding Production Systems.
\newblock
  \url{http://act-r.psy.cmu.edu/wordpress/wp-content/themes/ACT-R/tutorials/unit1.htm}.
\newblock Accessed: 2023-08-03.

\bibitem[{Anderson and Lebiere(2014)}]{anderson:2014}
Anderson, J.~R.; and Lebiere, C.~J. 2014.
\newblock \emph{The atomic components of thought}.
\newblock Psychology Press.

\bibitem[{Bender et~al.(2021)Bender, Gebru, McMillan-Major, and
  Shmitchell}]{bender:2021}
Bender, E.~M.; Gebru, T.; McMillan-Major, A.; and Shmitchell, S. 2021.
\newblock On the Dangers of Stochastic Parrots: Can Language Models Be Too Big?
\newblock In \emph{Proceedings of the 2021 ACM Conference on Fairness,
  Accountability, and Transparency}, FAccT '21, 610–623. New York, NY, USA:
  Association for Computing Machinery.
\newblock ISBN 9781450383097.

\bibitem[{Binz and Schulz(2023)}]{binz:2023}
Binz, M.; and Schulz, E. 2023.
\newblock Using cognitive psychology to understand GPT-3.
\newblock \emph{Proceedings of the National Academy of Sciences}, 120(6):
  e2218523120.

\bibitem[{Brown et~al.(2020)Brown, Mann, Ryder, Subbiah, Kaplan, Dhariwal,
  Neelakantan, Shyam, Sastry, Askell, Agarwal, Herbert-Voss, Krueger, Henighan,
  Child, Ramesh, Ziegler, Wu, Winter, Hesse, Chen, Sigler, Litwin, Gray, Chess,
  Clark, Berner, McCandlish, Radford, Sutskever, and Amodei}]{brown:2020}
Brown, T.; Mann, B.; Ryder, N.; Subbiah, M.; Kaplan, J.~D.; Dhariwal, P.;
  Neelakantan, A.; Shyam, P.; Sastry, G.; Askell, A.; Agarwal, S.;
  Herbert-Voss, A.; Krueger, G.; Henighan, T.; Child, R.; Ramesh, A.; Ziegler,
  D.; Wu, J.; Winter, C.; Hesse, C.; Chen, M.; Sigler, E.; Litwin, M.; Gray,
  S.; Chess, B.; Clark, J.; Berner, C.; McCandlish, S.; Radford, A.; Sutskever,
  I.; and Amodei, D. 2020.
\newblock Language Models are Few-Shot Learners.
\newblock In Larochelle, H.; Ranzato, M.; Hadsell, R.; Balcan, M.; and Lin, H.,
  eds., \emph{Advances in Neural Information Processing Systems}, volume~33,
  1877--1901. Curran Associates, Inc.

\bibitem[{Bubeck et~al.(2023)Bubeck, Chandrasekaran, Eldan, Gehrke, Horvitz,
  Kamar, Lee, Lee, Li, Lundberg, Nori, Palangi, Ribeiro, and
  Zhang}]{bubeck:2023}
Bubeck, S.; Chandrasekaran, V.; Eldan, R.; Gehrke, J.; Horvitz, E.; Kamar, E.;
  Lee, P.; Lee, Y.~T.; Li, Y.; Lundberg, S.; Nori, H.; Palangi, H.; Ribeiro,
  M.~T.; and Zhang, Y. 2023.
\newblock Sparks of Artificial General Intelligence: Early experiments with
  GPT-4.
\newblock arXiv:2303.12712.

\bibitem[{Caufield et~al.(2023)Caufield, Hegde, Emonet, Harris, Joachimiak,
  Matentzoglu, Kim, Moxon, Reese, Haendel et~al.}]{caufield:2023}
Caufield, J.~H.; Hegde, H.; Emonet, V.; Harris, N.~L.; Joachimiak, M.~P.;
  Matentzoglu, N.; Kim, H.; Moxon, S.~A.; Reese, J.~T.; Haendel, M.~A.; et~al.
  2023.
\newblock Structured prompt interrogation and recursive extraction of semantics
  (SPIRES): A method for populating knowledge bases using zero-shot learning.
\newblock \emph{arXiv preprint arXiv:2304.02711}.

\bibitem[{Devlin et~al.(2019)Devlin, Chang, Lee, and Toutanova}]{devlin:2019}
Devlin, J.; Chang, M.-W.; Lee, K.; and Toutanova, K. 2019.
\newblock {BERT}: Pre-training of Deep Bidirectional Transformers for Language
  Understanding.
\newblock In \emph{Proceedings of the 2019 Conference of the North {A}merican
  Chapter of the Association for Computational Linguistics: Human Language
  Technologies, Volume 1 (Long and Short Papers)}, 4171--4186. Minneapolis,
  Minnesota: Association for Computational Linguistics.

\bibitem[{Diao et~al.(2023)Diao, Wang, Lin, and Zhang}]{diao:2023}
Diao, S.; Wang, P.; Lin, Y.; and Zhang, T. 2023.
\newblock Active Prompting with Chain-of-Thought for Large Language Models.
\newblock arXiv:2302.12246.

\bibitem[{Du et~al.(2023)Du, Li, Torralba, Tenenbaum, and Mordatch}]{du:2023}
Du, Y.; Li, S.; Torralba, A.; Tenenbaum, J.~B.; and Mordatch, I. 2023.
\newblock Improving Factuality and Reasoning in Language Models through
  Multiagent Debate.
\newblock \emph{arXiv preprint arXiv:2305.14325}.

\bibitem[{Franklin and Patterson(2006)}]{franklin:2006}
Franklin, S.; and Patterson, F. 2006.
\newblock The LIDA architecture: Adding new modes of learning to an
  intelligent, autonomous, software agent.
\newblock \emph{pat}, 703: 764--1004.

\bibitem[{Gao et~al.(2023)Gao, Madaan, Zhou, Alon, Liu, Yang, Callan, and
  Neubig}]{gao:2023}
Gao, L.; Madaan, A.; Zhou, S.; Alon, U.; Liu, P.; Yang, Y.; Callan, J.; and
  Neubig, G. 2023.
\newblock PAL: Program-aided Language Models.
\newblock arXiv:2211.10435.

\bibitem[{Hesslow(2012)}]{hesslow:2012}
Hesslow, G. 2012.
\newblock The current status of the simulation theory of cognition.
\newblock \emph{Brain research}, 1428: 71--79.

\bibitem[{Huang et~al.(2023)Huang, Ruan, Huang, Jin, Dong, Wu, Bensalem, Mu,
  Qi, Zhao, Cai, Zhang, Wu, Xu, Wu, Freitas, and Mustafa}]{huang:2023}
Huang, X.; Ruan, W.; Huang, W.; Jin, G.; Dong, Y.; Wu, C.; Bensalem, S.; Mu,
  R.; Qi, Y.; Zhao, X.; Cai, K.; Zhang, Y.; Wu, S.; Xu, P.; Wu, D.; Freitas,
  A.; and Mustafa, M.~A. 2023.
\newblock A Survey of Safety and Trustworthiness of Large Language Models
  through the Lens of Verification and Validation.
\newblock arXiv:2305.11391.

\bibitem[{Kosinski(2023)}]{kosinski:2023}
Kosinski, M. 2023.
\newblock Theory of Mind May Have Spontaneously Emerged in Large Language
  Models.
\newblock arXiv:2302.02083.

\bibitem[{Kotseruba and Tsotsos(2020)}]{kotseruba:2020}
Kotseruba, I.; and Tsotsos, J.~K. 2020.
\newblock 40 years of cognitive architectures: core cognitive abilities and
  practical applications.
\newblock \emph{Artificial Intelligence Review}, 53(1): 17--94.

\bibitem[{Laird(2019)}]{laird:2019}
Laird, J.~E. 2019.
\newblock \emph{The Soar cognitive architecture}.
\newblock MIT press.

\bibitem[{Laird, Lebiere, and Rosenbloom(2017)}]{llr:2017}
Laird, J.~E.; Lebiere, C.; and Rosenbloom, P.~S. 2017.
\newblock A Standard Model of the Mind: Toward a Common Computational Framework
  across Artificial Intelligence, Cognitive Science, Neuroscience, and
  Robotics.
\newblock \emph{AI Magazine}, 38(4): 13--26.

\bibitem[{LeCun(2022)}]{lecun:2022}
LeCun, Y. 2022.
\newblock A path towards autonomous machine intelligence version 0.9. 2,
  2022-06-27.
\newblock \emph{Open Review}, 62.

\bibitem[{Li et~al.(2022)Li, Puig, Paxton, Du, Wang, Fan, Chen, Huang,
  Aky{\"{u}}rek, Anandkumar, Andreas, Mordatch, Torralba, and Zhu}]{li:2022}
Li, S.; Puig, X.; Paxton, C.; Du, Y.; Wang, C.; Fan, L.; Chen, T.; Huang, D.;
  Aky{\"{u}}rek, E.; Anandkumar, A.; Andreas, J.; Mordatch, I.; Torralba, A.;
  and Zhu, Y. 2022.
\newblock Pre-Trained Language Models for Interactive Decision-Making.
\newblock \emph{CoRR}, abs/2202.01771.

\bibitem[{Lieto, Lebiere, and Oltramari(2018)}]{lieto:2018}
Lieto, A.; Lebiere, C.; and Oltramari, A. 2018.
\newblock The knowledge level in cognitive architectures: Current limitations
  and possible developments.
\newblock \emph{Cognitive Systems Research}, 48: 39--55.
\newblock Cognitive Architectures for Artificial Minds.

\bibitem[{Marcus(2020)}]{marcus:2020}
Marcus, G. 2020.
\newblock The Next Decade in {AI:} Four Steps Towards Robust Artificial
  Intelligence.
\newblock \emph{CoRR}, abs/2002.06177.

\bibitem[{Mialon et~al.(2023)Mialon, Dessì, Lomeli, Nalmpantis, Pasunuru,
  Raileanu, Rozière, Schick, Dwivedi-Yu, Celikyilmaz, Grave, LeCun, and
  Scialom}]{mialon:2023}
Mialon, G.; Dessì, R.; Lomeli, M.; Nalmpantis, C.; Pasunuru, R.; Raileanu, R.;
  Rozière, B.; Schick, T.; Dwivedi-Yu, J.; Celikyilmaz, A.; Grave, E.; LeCun,
  Y.; and Scialom, T. 2023.
\newblock Augmented Language Models: a Survey.
\newblock arXiv:2302.07842.

\bibitem[{Minsky(1988)}]{minsky:1988}
Minsky, M. 1988.
\newblock \emph{Society of mind}.
\newblock Simon and Schuster.

\bibitem[{Park et~al.(2023)Park, O'Brien, Cai, Morris, Liang, and
  Bernstein}]{park:2023}
Park, J.~S.; O'Brien, J.~C.; Cai, C.~J.; Morris, M.~R.; Liang, P.; and
  Bernstein, M.~S. 2023.
\newblock Generative Agents: Interactive Simulacra of Human Behavior.
\newblock arXiv:2304.03442.

\bibitem[{Pavlick(2023)}]{pavlick:2023}
Pavlick, E. 2023.
\newblock Symbols and grounding in large language models.
\newblock \emph{Philosophical Transactions of the Royal Society A}, 381(2251):
  20220041.

\bibitem[{Qian et~al.(2022)Qian, Wang, Li, Li, and Yan}]{qian:2022}
Qian, J.; Wang, H.; Li, Z.; Li, S.; and Yan, X. 2022.
\newblock Limitations of language models in arithmetic and symbolic induction.
\newblock \emph{arXiv preprint arXiv:2208.05051}.

\bibitem[{Romero et~al.(2021)Romero, Wang, Zimmerman, Steinfeld, and
  Tomasic}]{romero:2021}
Romero, O.~J.; Wang, A.; Zimmerman, J.; Steinfeld, A.; and Tomasic, A. 2021.
\newblock A Task-Oriented Dialogue Architecture via Transformer Neural Language
  Models and Symbolic Injection.
\newblock In \emph{Proceedings of the 22nd Annual Meeting of the Special
  Interest Group on Discourse and Dialogue}, 438--444. Singapore and Online:
  Association for Computational Linguistics.

\bibitem[{Schick et~al.(2023)Schick, Dwivedi-Yu, Dessì, Raileanu, Lomeli,
  Zettlemoyer, Cancedda, and Scialom}]{schick:2023}
Schick, T.; Dwivedi-Yu, J.; Dessì, R.; Raileanu, R.; Lomeli, M.; Zettlemoyer,
  L.; Cancedda, N.; and Scialom, T. 2023.
\newblock Toolformer: Language Models Can Teach Themselves to Use Tools.
\newblock arXiv:2302.04761.

\bibitem[{Scialom~et al.(2022)}]{scialom:2022}
Scialom~et al., T. 2022.
\newblock Fine-tuned language models are continual learners.
\newblock In \emph{Proceedings of the 2022 Conference on Empirical Methods in
  Natural Language Processing}, 6107--6122.

\bibitem[{Shanahan(2006)}]{shanahan:2006}
Shanahan, M. 2006.
\newblock A cognitive architecture that combines internal simulation with a
  global workspace.
\newblock \emph{Consciousness and cognition}, 15(2): 433--449.

\bibitem[{Sun(2016)}]{sun:2016}
Sun, R. 2016.
\newblock \emph{Anatomy of the mind: exploring psychological mechanisms and
  processes with the Clarion cognitive architecture}.
\newblock Oxford University Press.

\bibitem[{Tomasic et~al.(2021)Tomasic, Romero, Zimmerman, and
  Steinfeld}]{tomasic:2021}
Tomasic, A.; Romero, O.~J.; Zimmerman, J.; and Steinfeld, A. 2021.
\newblock Propositional Reasoning via Neural Transformer Language Models.
\newblock \emph{Int. Workshop on Neural-Symbolic Learning and Reasoning
  (NESY)}.

\bibitem[{Venkit, Srinath, and Wilson(2022)}]{venkit:2022}
Venkit, P.~N.; Srinath, M.; and Wilson, S. 2022.
\newblock A Study of Implicit Bias in Pretrained Language Models against People
  with Disabilities.
\newblock In \emph{Proceedings of the 29th International Conference on
  Computational Linguistics}, 1324--1332. Gyeongju, Republic of Korea:
  International Committee on Computational Linguistics.

\bibitem[{Wang et~al.(2023)Wang, Xie, Jiang, Mandlekar, Xiao, Zhu, Fan, and
  Anandkumar}]{wang:2023}
Wang, G.; Xie, Y.; Jiang, Y.; Mandlekar, A.; Xiao, C.; Zhu, Y.; Fan, L.; and
  Anandkumar, A. 2023.
\newblock Voyager: An open-ended embodied agent with large language models.
\newblock \emph{arXiv preprint arXiv:2305.16291}.

\bibitem[{Wei et~al.(2022)Wei, Wang, Schuurmans, Bosma, Chi, Le, and
  Zhou}]{wei:2023}
Wei, J.; Wang, X.; Schuurmans, D.; Bosma, M.; Chi, E.~H.; Le, Q.; and Zhou, D.
  2022.
\newblock Chain of Thought Prompting Elicits Reasoning in Large Language
  Models.
\newblock \emph{CoRR}, abs/2201.11903.

\bibitem[{Weidinger et~al.(2022)Weidinger, Uesato, Rauh, Griffin, Huang,
  Mellor, Glaese, Cheng, Balle, Kasirzadeh, Biles, Brown, Kenton, Hawkins,
  Stepleton, Birhane, Hendricks, Rimell, Isaac, Haas, Legassick, Irving, and
  Gabriel}]{weidinger:2022}
Weidinger, L.; Uesato, J.; Rauh, M.; Griffin, C.; Huang, P.-S.; Mellor, J.;
  Glaese, A.; Cheng, M.; Balle, B.; Kasirzadeh, A.; Biles, C.; Brown, S.;
  Kenton, Z.; Hawkins, W.; Stepleton, T.; Birhane, A.; Hendricks, L.~A.;
  Rimell, L.; Isaac, W.; Haas, J.; Legassick, S.; Irving, G.; and Gabriel, I.
  2022.
\newblock Taxonomy of Risks Posed by Language Models.
\newblock In \emph{Proceedings of the 2022 ACM Conference on Fairness,
  Accountability, and Transparency}, FAccT '22, 214–229. New York, NY, USA:
  Association for Computing Machinery.
\newblock ISBN 9781450393522.

\bibitem[{Welleck et~al.(2019)Welleck, Kulikov, Roller, Dinan, Cho, and
  Weston}]{welleck:2019}
Welleck, S.; Kulikov, I.; Roller, S.; Dinan, E.; Cho, K.; and Weston, J. 2019.
\newblock Neural text generation with unlikelihood training.
\newblock \emph{arXiv preprint arXiv:1908.04319}.

\bibitem[{Xie et~al.(2023)Xie, Xie, Lin, Wei, Li, Kong, Chen, Zhuo, Hu, and
  Li}]{xie:2023}
Xie, Y.; Xie, T.; Lin, M.; Wei, W.; Li, C.; Kong, B.; Chen, L.; Zhuo, C.; Hu,
  B.; and Li, Z. 2023.
\newblock OlaGPT: Empowering LLMs With Human-like Problem-Solving Abilities.
\newblock arXiv:2305.16334.

\bibitem[{Yao et~al.(2023{\natexlab{a}})Yao, Yu, Zhao, Shafran, Griffiths, Cao,
  and Narasimhan}]{yao:2023tree}
Yao, S.; Yu, D.; Zhao, J.; Shafran, I.; Griffiths, T.~L.; Cao, Y.; and
  Narasimhan, K. 2023{\natexlab{a}}.
\newblock Tree of Thoughts: Deliberate Problem Solving with Large Language
  Models.
\newblock arXiv:2305.10601.

\bibitem[{Yao et~al.(2023{\natexlab{b}})Yao, Zhao, Yu, Du, Shafran, Narasimhan,
  and Cao}]{yao:2023}
Yao, S.; Zhao, J.; Yu, D.; Du, N.; Shafran, I.; Narasimhan, K.; and Cao, Y.
  2023{\natexlab{b}}.
\newblock ReAct: Synergizing Reasoning and Acting in Language Models.
\newblock arXiv:2210.03629.

\bibitem[{Zhang et~al.(2022)Zhang, Zhang, Li, and Smola}]{zhang:2022}
Zhang, Z.; Zhang, A.; Li, M.; and Smola, A. 2022.
\newblock Automatic Chain of Thought Prompting in Large Language Models.
\newblock arXiv:2210.03493.

\bibitem[{Zhu et~al.(2023)Zhu, Chen, Shen, Li, and Elhoseiny}]{zhu:2023}
Zhu, D.; Chen, J.; Shen, X.; Li, X.; and Elhoseiny, M. 2023.
\newblock Minigpt-4: Enhancing vision-language understanding with advanced
  large language models.
\newblock \emph{arXiv preprint arXiv:2304.10592}.

\end{thebibliography}

\end{document}